# Fronto-parietal and fronto-temporal EEG coherence as predictive neuromarkers of transcutaneous auricular vagus nerve stimulation response in treatment-resistant schizophrenia: A machine learning study

Yapeng Cui[1]; Ruoxi Yun[2]; Shumin Zhang[4]; Yi Gong[3]; Zhiqin Li[3]; Ying Chen[3]; Mingbing Su[3]; Dongniya Wu[3]; Jingxia Wu[3]; Qian Wang[3]; Jianan Wang[2]; Qianqian Tian[2]; Yangyang Yuan[1]; Shuhao Mei[2]; Lei Wu[3]; Xinghua Li[3]; Bingkui Zhang[3,✉]; Taipin Guo[3,4,✉]; Jinbo Sun[1,2,✉]

[1]Engineering Research Center of Molecular and Neuro Imaging of the Ministry of Education, School of Life Science and Technology, Xidian University, Xian, 710126, China. [2]Wearable BCI and Intelligent Rehabilitation Innovation Lab, Guangzhou Institute of Technology, Xidian University, Guangzhou, 510530, China. [3]Kunming Mental Hospital, Kunming, 650000, China. [4]School of Second Clinical Medicine/The Second Affiliated Hospital, Yunnan University of Chinese Medicine, Kunming, 650500, China.

✉Corresponding authors:

Jinbo Sun: No.266 Xinglong Section of Xifeng Road, Xi'an City, Shaanxi Province, China; +8618629557811; sunjb@xidian.edu.cn.

Bingkui Zhang: Cold Water Pond, Xizhu Sub District Office, Taoyuan Community, Wuhua District, Kunming City, Yunnan Province, China; +8613700690166; 627781482@qq.com.

Taipin Guo: No.1076 Yuhua Road, Xincheng Subdistrict, Chenggong District, Kunming City, Yunnan Province, China; +8618487272658; gtphncs@126.com.

# Abstract

**Background:** Response variability limits the clinical utility of transcutaneous auricular vagus nerve stimulation (taVNS) for negative symptoms in treatment-resistant schizophrenia (TRS). This study aimed to develop an electroencephalography (EEG)-based machine learning (ML) model to predict individual response and explore associated neurophysiological mechanisms.

**Methods:** We used ML to develop and validate predictive models based on pre-treatment EEG data features (power, coherence, and dynamic functional connectivity) from 50 TRS patients enrolled in the taVNS trial (http://www.chictr.org, ChiCTR2400085198), within a nested cross-validation framework. Participants received 20 sessions of active or sham taVNS (n = 25 each) over two weeks, followed by a two-week follow-up. The prediction target was the percentage change in the positive and negative syndrome scale-factor score for negative symptoms (PANSS-FSNS) from baseline to post- treatment, with further evaluation of model specificity and neurophysiological relevance.

**Results:** The optimal model accurately predicted taVNS response in the active group, with predicted PANSS-FSNS changes strongly correlated with observed changes ($r = 0.87$, $p < .001$); permutation testing confirmed performance above chance ($p < .001$). Nine consistently retained features were identified, predominantly fronto-parietal and fronto-temporal coherence features. Negligible predictive performance in the sham group and failure to predict positive symptom change support the predictive specificity of this oscillatory signature for taVNS-related negative symptom improvement. Two coherence features within fronto-parietal-temporal networks showed post-taVNS changes significantly associated with symptom improvement, suggesting dual roles as predictors and potential therapeutic targets.

**Conclusion:** EEG oscillatory neuromarkers enable accurate prediction of individual taVNS response in TRS, supporting mechanism-informed precision neuromodulation strategies.

# 1. Introduction

The treatment of negative symptoms in schizophrenia remains a major unmet need, as antipsychotics show limited efficacy(Fusar-Poli et al., 2015; McCutcheon et al., 2020). These symptoms are prevalent in treatment-resistant schizophrenia (TRS), contributing to high disease burden and poor prognosis(Galderisi et al., 2018; Owen et al., 2016; Sicras-Mainar et al., 2014). We recently demonstrated, for the first time, the efficacy and safety of transcutaneous auricular vagus nerve stimulation (taVNS) for negative symptoms in TRS(Cui et al., 2025). However, as with other non-invasive brain stimulation (NIBS) methods, response was variable, with only 44% achieving clinical response. This variability in treatment response may be attributable to inter-individual differences in neuroanatomical and neurofunctional(Koutsouleris et al., 2018; Tu et al., 2018), underscoring the need to identify reliable predictors to guide personalized and optimized treatment.

Machine learning (ML) has emerged as an important approach for predicting treatment outcomes in schizophrenia, particularly due to its advantages in handling small clinical sample sizes(Iniesta et al., 2016; Sharaev et al., 2022). Within this framework, neurobiological features derived from magnetic resonance imaging (MRI) and electroencephalography (EEG) have been increasingly used to predict individual treatment response. MRI-based ML models have shown promising performance in predicting antipsychotic treatment response. For example, Chen et al. developed models using white matter tracts, gray matter volume, and cortical thickness, enabling individualized prediction of antipsychotic response in schizophrenia(Chen et al., 2023; Y. Chen et al., 2024). Beyond pharmacotherapy, MRI-based ML approaches have been applied to predict responses to NIBS techniques, including electroconvulsive therapy and repetitive transcranial magnetic stimulation, with high classification accuracy(Dong et al., 2024; Koutsouleris et al., 2018; Li et al., 2017; Xi et al., 2020) and moderate predictive correlations(Gong et al., 2020; Yang et al., 2020). In parallel, EEG-based ML approaches have demonstrated strong predictive value for antipsychotic outcomes, especially in TRS. Ciprian et al. integrated diverse EEG features with ML, achieving prediction accuracies of 89.90% and 95.83% for clozapine response in TRS(Ciprian et al., 2020; Masychev et al., 2020). Although EEG-based prediction studies remain relatively limited, existing reviews suggest EEG may provide higher sensitivity and specificity than other approaches(Saboori Amleshi et al., 2025). Given its low cost, convenience, and clinical accessibility(Zhang et al., 2023), EEG represents an attractive modality for precision prediction in real-world settings.

However, despite encouraging findings in pharmacological response prediction, few studies have employed EEG to predict the efficacy of NIBS techniques in schizophrenia. This gap is clinically relevant, because EEG abnormalities are closely tied to schizophrenia pathophysiology and symptom expression(Ahmad et al., 2022; Hasey and Kiang, 2013; Itil, 1977). Alpha- and theta-band power have been validated as diagnostic biomarkers for schizophrenia across multicenter datasets(Jiang et al., 2023), while reduced EEG phase coherence and abnormal interregional connectivity reflect impairments in temporal integration, synaptic plasticity, and network function in patients(Mackintosh et al., 2021; Stephan et al., 2006; Wolff and Northoff, 2024). Importantly, for negative symptoms poorly responsive to antipsychotics, specific neurophysiological features, including frontal abnormalities, dysregulated alpha/theta/beta oscillations, and reduced EEG coherence, have been linked to greater negative symptom severity(Gerez and Tello, 1995; Hudgens-Haney et al., 2020; Sharaev et al., 2022; Shukla et al., 2019). Together, these findings provide a basis for selecting EEG features linked to negative symptoms and building predictive models of taVNS efficacy.

Given the demonstrated predictive value of EEG for treatment response, its favorable clinical accessibility, and our preliminary findings linking EEG feature changes to taVNS-induced improvements in negative symptoms in TRS(Cui et al., 2025), the present study aims to integrate EEG features with ML to predict taVNS treatment outcomes in TRS. Based on data from the first randomized controlled trial of taVNS for negative symptoms in TRS(Cui et al., 2025), we developed a nested cross-validation ML framework with EEG power, coherence, and dynamic functional connectivity features to predict taVNS response. Furthermore, we examined the specificity of the prediction model for taVNS-induced improvements in negative symptoms and its temporal stability during follow-up, and explored EEG features potentially associated with treatment response.

## 2. Methods

### 2.1 Study subjects and endpoints for individualized prediction

We included inpatients who were diagnosed with schizophrenia by psychiatrists according to the ICD-10 criteria and who met the definition of TRS. Participants were between 18 and 65 years old, had received a stable dose of antipsychotic medication for at least two months prior to screening, and maintained a stable dose throughout the study. They had not experienced an acute exacerbation of positive symptoms within the six months before screening, had a score of 24 or higher on the PANSS-factor score for negative symptoms (PANSS-FSNS), and scored 4 or higher on at least two of the three negative PANSS items during both screening and baseline (N1 blunted affect, N4 passive/apathetic social withdrawal, N6 lack of spontaneity and flow of conversation)(Leucht et al., 2019). The main exclusion criteria included prominent positive or depressive symptoms, neurological disorders, comorbid psychiatric conditions, or other factors that could interfere with the study or compromise data integrity (Supplementary Methods). The study was approved by the Ethics Committee of the Kunming Psychiatric Hospital, China, and registered in the Chinese Clinical Trial Registry (ChiCTR2400085198) on June 3, 2024(Cui et al., 2025). In accordance with the Declaration of Helsinki, all participants provided written informed consent prior to enrollment.

A total of 50 patients underwent pre-treatment EEG acquisition (active taVNS/sham taVNS: N = 25/25), followed by a two-week taVNS intervention and a two-week follow-up period (T0 to T4), with clinical assessments conducted at the end of each week. The primary outcome was the change in PANSS-FSNS scores from baseline to post-intervention, representing the remission rate of negative symptoms(Obermeier et al., 2010), which was calculated as:

$$(PANSS-FSNS)(\%) = \frac{(PANSS-FSNS)_{T0} - (PANSS-FSNS)_{T1}}{(PANSS-FSNS)_{T0} - 7}$$

This percentage reduction was used as the target variable for ML prediction.

## 2.2 Intervention

Participants underwent a total of 20 sessions of active or sham taVNS intervention over two weeks, administered five days per week, twice daily, with each session lasting 30 minutes. The taVNS device used in this study is an improved version of the model used in our previous research (TES-M3-M-01, Wearable Brain-Computer Interface and Intelligent Rehabilitation Technology Innovation Laboratory, Xidian University, Guangzhou Institute of Technology,

China)(Shen et al., 2022; Shi et al., 2024; Sun et al., 2021b, 2021a; Tian et al., 2024, 2023; Zhao et al., 2023, 2022).

The taVNS stimulation electrodes were placed on the left cymba conchae and tragus, delivering currents in the range of 0.1 to 3 mA, adjusted in steps of 0.1 mA. The stimulation intensity was set at the midpoint between the participant's sensory and pain thresholds, determined by repeated measurements. The taVNS pulse frequency was 25 Hz(Evensen et al., 2022; Kaczmarczyk et al., 2021), with a pulse width of 250 μs, and a 30-second on-off cycle(Frangos et al., 2015a). For active taVNS, the current intensity increased at 0.1 mA/s to the target level within 30 seconds, maintained for approximately 29 minutes, and then decreased at the same rate. For sham taVNS, stimulation was delivered only at the beginning and end, with current rising and falling at 0.1 mA/s(Frangos et al., 2015b; Tian et al., 2023). The stimulation site was identical in both conditions. Participants were informed they might or might not feel sensations during the intervention.

As previously reported, our intention-to-treat analysis demonstrated that a two-week taVNS intervention effectively alleviated negative symptoms in TRS and was well tolerated. At the end of treatment, the primary outcome measure, the PANSS-FSNS, showed significantly greater improvement in the active taVNS group compared with the sham group, and this effect persisted throughout the follow-up period(Cui et al., 2025).

## 2.3 EEG data acquisition and processing

EEG was recorded from 16 scalp electrodes (Fp1, Fp2, F7, F3, F4, F8, C3, C4, P3, P4, T3, T4, T5, T6, O1, O2) according to the international 10-20 system during approximately 5 minutes of eyes-closed resting state at baseline and post-treatment (Neurofax EEG-1200, Nihon Kohden, Japan). Electrode impedance was kept below 10 kΩ, and the sampling rate was 500 Hz. All EEG data were preprocessed in MATLAB using EEGLAB(Delorme and Makeig, 2004), and EEG power, coherence, and dynamic functional connectivity(H. Chen et al., 2024) features were subsequently calculated, yielding a total of 912 features (Supplementary Methods).

## 2.4 Establishment of a predictive model

We employed a support vector regression (SVR) model(Vapnik et al., n.d.) to predict the efficacy of taVNS in patients with TRS. Given the modest sample size, model performance was evaluated using a nested cross-validation framework to reduce optimism bias and improve the robustness of internal performance estimation (Varma and Simon, 2006). In the outer loop, a five-fold cross-validation was applied to assess the generalizability of the model to unseen data, while the inner loop simultaneously performed feature engineering and hyperparameter optimization to prevent data leakage (Supplementary Fig. S1).

**Data preprocessing.** Prior to model training and feature selection, systematic preprocessing procedures were applied to ensure data quality and robustness, including missing value imputation, outlier correction, and normalization (Supplementary Methods).

**Feature selection.** To address the issue of high-dimensional input and multicollinearity(Guyon and Elisseeff, n.d.; Zou and Hastie, 2005), we adopted an elastic net-based feature selection method. This method combines the sparsity of L1 regularization (Lasso) with the stability of L2 regularization (Ridge)(Hoerl and Kennard, n.d.; Tibshirani, 1996). We first performed five-fold cross-validation on the feature set and independently applied elastic net regression on the training set of each fold to select the most predictive features. In this process, to further assess the stability and robustness of feature selection, leave-one-out cross-validation (LOOCV) was applied within each training fold(Fan et al., 2011). During training, the indices of features selected by the elastic net model in each iteration were recorded. Features selected in at least three out of five five folds were retained as the final subset (Supplementary Methods).

**Prediction model establishment.** After defining the preliminary feature subset, forward stepwise regression(Bendel and Afifi, 1977) was applied to iteratively add features and evaluate their predictive contribution using a linear-kernel SVR. To enhance stability and generalizability, we implemented a nested cross-validation framework(Varma and Simon, 2006): LOOCV was embedded within five-fold cross-validation, and the mean Pearson correlation coefficient ($r$) across folds quantified performance. Model hyperparameters ($C$ ($10^{-2}$-$10^{3}$, logarithmic scale) and $\varepsilon$ (0.01-1.0, linear scale))(Bergstra and Bengio, n.d.) were optimized via grid search within the inner folds to prevent data leakage. The feature subset yielding the highest mean r-value was chosen as the final set and validated in the outer folds to confirm predictive reliability.

**Prediction model evaluation.** After determining the optimal feature subset, model performance was evaluated on the outer cross-validation test sets. Specifically, the trained SVR model was applied to predict PANSS-FSNS(%) scores for each subject in the test set, and predictive accuracy was quantified by calculating the correlation between predicted and observed scores, as well as error metrics including root mean squared error (RMSE) and mean absolute error (MAE). Furthermore, to examine the statistical significance of the predictive performance, we conducted a permutation test by randomly shuffling the outcome labels and repeating the entire SVR training and cross-validation procedure 1,000 times, thereby generating a null distribution against which the observed performance was compared(Golland and Fischl, 2003) (Supplementary Fig. S2).

## 2.5 Post hoc analyses

We conducted a series of post hoc analyses using R (version 4.2.3). (1) Participants in both the active and sham groups were classified into predicted responders and non-responders based on whether the model-predicted PANSS-FSNS(%) exceeded 20%(Howes et al., 2017; Wobrock et al., 2015). Subsequently, a linear mixed-effects model (nlme package) was constructed with PANSS-FSNS change from baseline as the dependent variable, with time (T2, T3, T4), group (predicted responder vs. non-responder), and their interaction as fixed effects, and participants as random intercepts (or participants as random intercepts with time effect as random slopes). Model selection was performed using likelihood ratio tests to identify the better-fitting model, thereby examining whether responder classification at treatment endpoint could capture the sustained effects of taVNS over the follow-up period. (2) Furthermore, for the features selected by the predictive model, we combined them with post-treatment EEG data to examine correlations between feature changes during treatment and PANSS-FSNS(%). This analysis aimed to explore whether the predictive features may also serve as potential biomarkers of negative symptom improvement. Correlation analysis were computed using a permutation-based Spearman rank correlation method (100,000 permutations)(Pesarin and Salmaso, 2010), which allows for quantification of the monotonic relationship between two variables without assuming linearity or normal distribution of the data.

# 3. Results

## 3.1 Sociodemographic and clinical efficacy

At baseline, there were no significant differences in the PANSS-FSNS and other demographic and clinical characteristics between the two groups (Table 1). In our previous intention-to-treat analysis, patients receiving active taVNS showed a significantly greater improvement in negative symptoms compared with those receiving the sham procedure (PANSS-FSNS difference, -1.36; effect size, -0.62, 95% CI, -1.20 to -0.04; $p = 0.033$), with effects sustained at follow-up and good tolerability. Moreover, the distribution of PANSS-FSNS responders and non-responders differed significantly between active and sham groups (taVNS responders/non-responders vs. sham responders/non-responders: 11/14 vs. 2/25; $p = 0.008$)(Cui et al., 2025).

## 3.2 Prediction of treatment outcome

In the active taVNS group, EEG power spectra, coherence, and dynamic functional connectivity features were employed as predictors to predict post-treatment PANSS-FSNS(%) scores. During the five-fold elastic net feature selection procedure, correlations between baseline features and treatment response were calculated within each training set. A total of 25 common EEG features were identified, consistently selected in at least three of the five folds (Supplementary Fig. S3). Based on the 25 selected features, the nested cross-validation SVR model with stepwise regression demonstrated a significant correlation between the predicted and observed PANSS-FSNS(%) scores ($r = 0.87$, $p < .001$) (Table 2; Fig. 1A). In the final predictive model of PANSS-FSNS(%), 9 EEG coherence features were retained (Fig. 1B). Importantly, to avoid potential randomness introduced by cross-validation, a 1,000-iteration permutation test confirmed that the correlation between predicted and observed PANSS-FSNS(%) scores derived from the nested cross-validated SVR model was statistically significant ($p < .001$; Supplementary Fig. S2).

When assessing whether the predictive model was specific to taVNS and the improvement of negative symptoms in patients with TRS, we found that, the predictive pattern of PANSS-FSNS(%) specifically yielded robust prediction accuracy in the active, but not in the sham group (RMSE = 0.11, MAE = 0.09, $r = -0.36$, $p = 0.079$; Table 2; Fig. 2A); similarly, our the model pipeline produced an accurate predictor for the PANSS-FSNS(%) endpoint in the active group, but not for PANSS-FSPS(%) endpoint (RMSE = 0.39, MAE = 0.30, $r = 0.14$, $p = 0.495$; Table 2; Fig. 2B).

## 3.3 Long-term stability of prediction results

The mixed-model analysis applied to the relative change PANSS-FSNS scores of this patient group revealed a significant main effect of the predicted outcome group on the PANSS-FSNS trajectories ($F = 7.31$, $p < .001$; Supplementary Table S1; Fig. 2C). The trajectories derived from the mixed-effects model included PANSS-FSNS observations at the intervention endpoint (T2) and follow-up assessments (T3 and T4). These analyses demonstrated that the stratification between predicted responders and non-responders remained stable two weeks after taVNS intervention, whereas no such effect was observed in the sham group. Additional mixed-model analyses showed that patients stratified based on predicted response status also differed significantly in their PANSS negative subscale scores trajectories ($F = 4.00$, $p = 0.011$). However, no comparable long-term stratification stability was observed for PANSS-FSPS or PANSS positive subscale scores throughout the course (Supplementary Table S1).

### 3.4 Efficacy related outcomes for selected features

Among the 9 EEG features selected by the final PANSS-FSNS predictive model in the active group, beta-band coherence between F3 and P3 was positively correlated with PANSS-FSNS improvement ($r_{F3\_P3\_beta}$ = 0.66, $p_{F3\_P3\_beta}$ <.001), whereas gamma-band coherence between P4 and T3 were negatively correlated ($r_{P4\_T3\_gamma}$ = -0.47, $p_{P4\_T3\_gamma}$ = 0.021). These associations were not observed in the sham group (Fig. 3).

## 4. Discussion

To our knowledge, this is the first study to apply an EEG oscillation-based machine-learning approach to predict individual response to taVNS in patients with TRS. While our prior work (Cui et al., 2025) demonstrated the efficacy of taVNS for negative symptoms in TRS, the present study addresses a complementary question: which patients are most likely to benefit from this treatment? This shift from demonstrating efficacy to predicting individual treatment response represents an important step toward precision psychiatry. Using data from a randomized controlled trial, we implemented a nested cross-validation SVR model that accurately predicted negative symptom improvement, with predicted and observed PANSS-FSNS(%) showing a strong correlation ($r$ = 0.87, $p$ < .001) and permutation testing confirming performance above chance ($p$ < .001). Nine EEG features were consistently selected, predominantly coherence features, predominantly fronto-parietal and fronto-temporal coherence features, and the model showed specificity for taVNS-related negative symptom improvement, with predictive performance remaining stable during follow-up. Moreover, two EEG features showed intervention-related changes associated with symptom improvement, suggesting potential roles as both predictive markers and candidate neural targets of taVNS.

Validation analyses indicated that the taVNS response prediction model demonstrated not only high predictive accuracy but also clear treatment specificity. The markedly reduced predictive performance in patients receiving sham taVNS supports its high specificity for active taVNS-induced improvement in negative symptoms. Moreover, the model achieved significant predictive accuracy for PANSS-FSNS(%) but showed substantially weaker performance for PANSS-FSPS(%) in the active taVNS group, further reinforcing the specificity of the identified neurophysiological features to taVNS-related neural effects underlying negative symptom improvement. This intervention specificity is consistent with our previous findings(Cui et al., 2025) showing that taVNS selectively alleviates negative symptoms in TRS with sustained effects during follow-up, potentially through modulation of cortical oscillatory activity in brain regions closely implicated in negative symptoms, including frontal and parietal areas.

The nine EEG coherence features demonstrated high discriminative power in predicting taVNS-induced improvement in negative symptoms, spanning multiple brain regions and frequency bands. This finding is consistent with the view that core neurobiological deficits in schizophrenia are not restricted to specific regions, but rather arise from dysfunctional interactions within and between cortical and subcortical networks(Uhlhaas and Singer, 2015, 2012). The selected features covered a broad frequency range, including low-frequency delta as well as high-frequency beta and gamma bands. Previous studies have reported abnormal EEG coherence across tasks and frequencies in schizophrenia(Wolff and Northoff, 2024), and a review by Hudgens-Haney et al. further linked resting-state multiband EEG features to negative symptoms(Hudgens-Haney et al., 2020), providing converging support for our results.

From a spatial perspective, most of the nine features were concentrated in connections between the frontal lobe and other cortical regions, suggesting that abnormalities in frontal connectivity may be a key factor influencing treatment response. Previous studies have shown that the frontal lobe, particularly the dorsolateral prefrontal cortex, plays a central role in motivation regulation, executive control, and socio-emotional processing, and is widely implicated in negative symptoms(Andreasen et al., 1997; Shukla et al., 2019). The connectivity features included in the predictive model, including fronto-parietal connections (F3-P3, Fp1-C4, Fp1-C3, C3-P3) and fronto-temporal connections (F4-F7, Fp2-F7), involve the fronto-parietal-temporal network, where coherence abnormalities have been widely reported in schizophrenia(Chang et al., 2024; Prieto-Alcántara et al., 2023; Wolff and Northoff, 2024). In addition, these coherence features exhibited an overall left-lateralized spatial distribution and a predominance of long-range connectivity. Previous studies have demonstrated abnormal hemispheric asymmetry in schizophrenia(Oertel et al., 2010; Royer et al., 2015), and both neurophysiological and structural asymmetry indices have been associated with the severity of negative symptoms(Bleich-Cohen et al., 2012; Huang et al., 2022; Mitra et al., 2017). Moreover, long-range synchrony may depend on excitatory pathways implicated in the excitation/inhibition (E/I) imbalance hypothesis of schizophrenia (Engel et al., 1991; Vicente et al., 2008) and may also involve long-range GABAergic connectivity(Lee et al., 2014; Melzer et al., 2012), both of which have been considered important targets for alleviating negative symptoms(Uliana et al., 2024). These features may therefore reflect key network-level dysfunctions related to negative symptoms and further contribute to individual differences in response to taVNS. Taken together, the above evidence supports the notion that nine multiband, predominantly frontal, left-lateralized, and long-range EEG coherence features may serve as robust neuro-oscillatory markers for predicting taVNS-induced improvement in negative symptoms in TRS. Their predictive value may arise from sensitive representation of the functional state of neural networks associated with negative symptoms。

Importantly, we found that changes in two predictive coherence features, F3-P3-beta and P4-T3-gamma, during taVNS treatment were significantly associated with improvement in negative symptoms, whereas no such associations were observed in the sham taVNS group. This finding suggests that these features may function not only as predictors of taVNS-induced improvement in negative symptoms, but also as potential therapeutic targets. Previous studies have reported abnormalities in functional connectivity and neural oscillations in TRS(Molent et al., 2019; Nucifora et al., 2019), particularly attenuated connectivity in higher-frequency bands such as beta and gamma(Ogyu et al., 2023; Takahashi et al., 2018). In addition, taVNS has been shown to modulate brain oscillatory activity across multiple frequency bands and alter connectivity among distributed brain regions, especially within neural circuits involved in emotional and interoceptive regulation(Gianlorenco et al., 2022; Keatch et al., 2023; Zhang et al., 2024). Neuroimaging evidence further indicates that taVNS engages ascending vagal pathways projecting to the nucleus tractus solitarius, subsequently influencing activity and connectivity in emotion- and interoception-related subcortical regions, including the insula, hypothalamus, and amygdala(Engelen et al., 2023; Ferstl et al., 2022; Frangos et al., 2015a; Keatch et al., 2023; Poppa et al., 2022; Ventura-Bort and Weymar, 2024; Zhang et al., 2024), which closely interact with cortical regions such as the frontal, parietal, and temporal lobes. Taken together with the observed association between changes in fronto-parietal-temporal coherence features and negative symptom improvement, we reasonably speculate that taVNS may modulate oscillatory synchronization across interacting cortical and subcortical networks, thereby acting on functional circuits relevant to negative symptoms and influencing clinical outcomes in TRS. Notably, longitudinal changes in these two EEG coherence features were observed exclusively in the active taVNS group and not in the sham group, further supporting their involvement in taVNS-specific effects. Accordingly, these features may not only reflect state-dependent changes in brain activity following taVNS, but also represent potential neural mechanisms underlying its effects on negative symptoms, with future utility for monitoring treatment response, guiding parameter optimization, and informing individualized taVNS interventions through neurophysiological feedback.

The feasibility of predicting taVNS treatment response from neural oscillatory features is strengthened by the randomized controlled trial design, which enabled evaluation of intervention specificity (active vs. sham), outcome-domain specificity, and the temporal stability of model predictions. Given the substantial heterogeneity in taVNS response that has limited its broader clinical application, our model may help identify likely responders before treatment. In addition, EEG features associated with treatment response highlight candidate neural network targets of taVNS, providing a physiological basis for further mechanistic investigation and the future development of closed-loop treatment strategies.

Nevertheless, several limitations should be acknowledged. First, the sample size was relatively small for a machine-learning study focused on individual-level prediction. To mitigate this limitation, we employed nested cross-validation, which separates model development from performance estimation and helps reduce overfitting and optimism bias in small-sample predictive modeling. In addition, permutation testing and sham-group specificity analyses provided further support for the robustness of the observed predictive signal. Second, the dataset was derived from a single-center cohort, limiting direct assessment of generalizability beyond the present study setting. Accordingly, these findings should be considered preliminary and require replication and external validation in larger, multicenter samples. Nevertheless, they provide proof of concept that EEG coherence features may serve as candidate predictors of taVNS response in TRS. In the context of rare and difficult-to-treat conditions such as TRS, even modest-sized but methodologically rigorous studies can offer useful guidance for future trials.

## 5. Conclusion

In summary, we developed a machine-learning model based on pre-treatment neural oscillatory features that accurately predicted taVNS-related improvement in negative symptoms in patients with TRS. These findings support the potential of EEG-based markers to improve patient stratification and reduce response heterogeneity in taVNS treatment. In addition, several predictive EEG features may also reflect neurophysiological processes underlying symptom improvement, providing a basis for future work to optimize stimulation protocols, monitor treatment response, and develop individualized neuromodulation strategies.

## Data availability

Study-related data are available from the corresponding author upon reasonable request and with approval from the hospital administration.

## Acknowledgements

This work was supported by the National Natural Science Foundation of China (grant number 82441050), Xidian University Specially Funded Project for Interdisciplinary Exploration (grant number TZJH2024014), Natural Science Basic Research Program of Shaanxi (grant number 2025SYS-SYSZD-061), Science and Technology Plan Project of the Department of Science and Technology of Yunnan Province (grant number 20230IAZ070001-121) and Yunnan High-level Talents in Traditional Chinese Medicine Discipline Leader (Chinese Medicine Acupuncture; Taipin Guo), which is gratefully acknowledged. The funders had no role in the study design, data collection and analysis, decision to publish, or preparation of the manuscript. Although some of the authors are affiliated with Xidian University, the university had no influence on the study results or interpretation. We thank all research assistants, physicians, and nursing staffs at Kunming Psychiatric Hospital for their assistance during the study process, as well as the technicians in the EEG room, who performed the EEG collection, without whom this work could have been possible. We are also grateful to Yixuan Wu, Yu Tao and Hao Jing for their valuable support during the early stages of the project.

## Author contributions

**Yapeng Cui:** Conceptualization, Data curation, Formal analysis, Methodology, Writing–review

& editing, Writing–original draft. **Ruoxi Yun:** Formal analysis, Methodology, Software, Validation, Visualization. **Shumin Zhang:** Data curation, Methodology, Validation. **Yi Gong:** Project administration, Resources. **Zhiqin Li:** Investigation, Resources. **Ying Chen:** Investigation, Resources. **Mingbing Su:** Investigation, Resources. **Dongniya Wu:** Investigation, Resources. **Jingxia Wu:** Investigation, Resources. **Qian Wang:** Investigation, Resources. **Jianan Wang:** Data curation. **Qianqian Tian:** Software. **Yangyang Yuan:** Data curation. **Shuhao Mei:** Software. **Lei Wu:** Resources. **Xinghua Li:** Resources. **Bingkui Zhang:** Conceptualization, Funding acquisition, Project administration, Resources. **Taipin Guo:** Conceptualization, Funding acquisition. **Jinbo Sun:** Conceptualization, Funding acquisition, Project administration, Writing–review & editing.

## Competing interests

The authors declare no competing interests.

## Ethics approval

This study was performed in accordance with the Declaration of Helsinki and was approved by the Kunming Psychiatric Hospital in China Review Board. Informed consent was provided by all of the participants.

## AI Statement

Portions of this manuscript were refined using an AI-based language tool (ChatGPT) to improve grammar, clarity, and readability. The authors reviewed and approved all content, and the AI tool was not used to generate scientific ideas, analysis, or conclusions.

# Figure and table

**Fig. 1.** Nine EEG features selected by the optimal model and prediction results of taVNS treatment efficacy in patients with TRS.

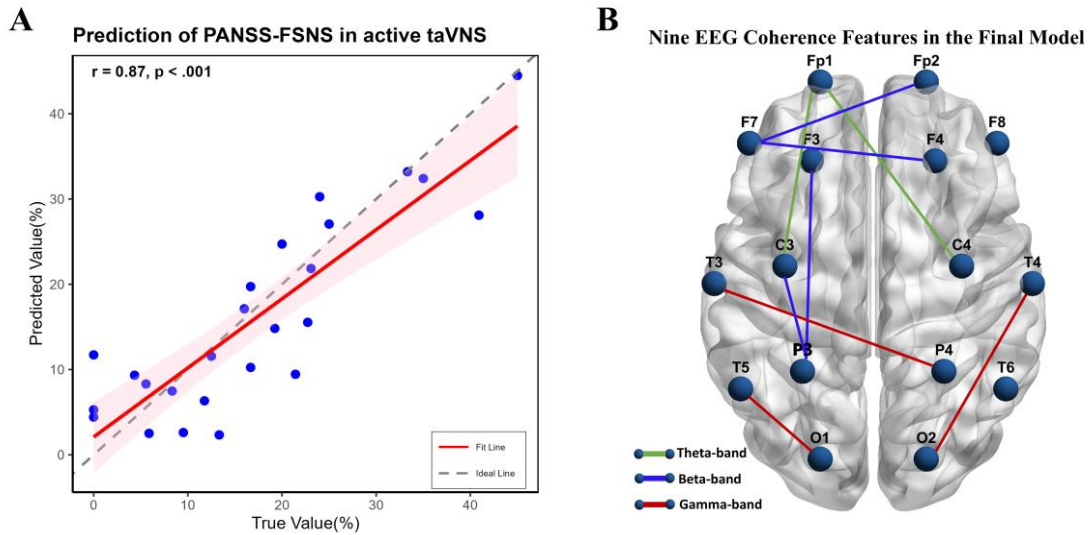

(A) Nine EEG coherence features in the optimal model. (B) Individual-level correlation between predicted and observed PANSS-FSNS negative symptom improvement following treatment in the active taVNS group.

**Fig. 2.** The specificity of the prediction model for taVNS-induced improvements in negative symptoms and its temporal stability during follow-up.

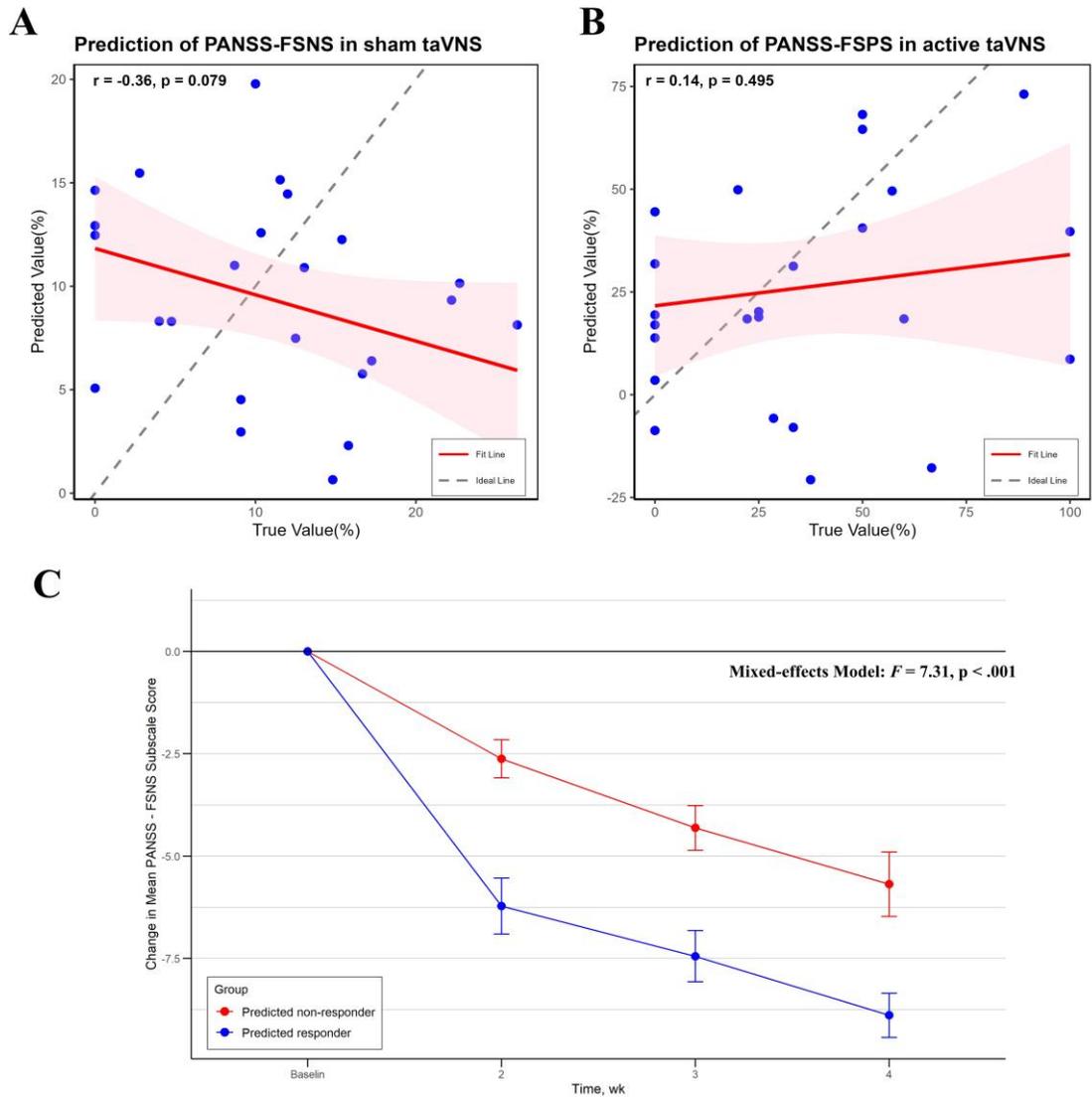

(A) Individual-level correlation between predicted and observed PANSS-FSNS negative symptom improvement following treatment in the sham taVNS group. (B) Individual-level correlation between predicted and observed PANSS-FSPS positive symptom improvement following treatment in the active taVNS group. (C) Descriptive trajectory graphs of PANSS-FSNS changes in patients with predicted nonresponse (red) and response (blue) to active taVNS, spanning from baseline to the end of the second-week follow-up.

**Fig. 3.** Correlation between selected EEG features and PANSS-FSNS improvement rate.

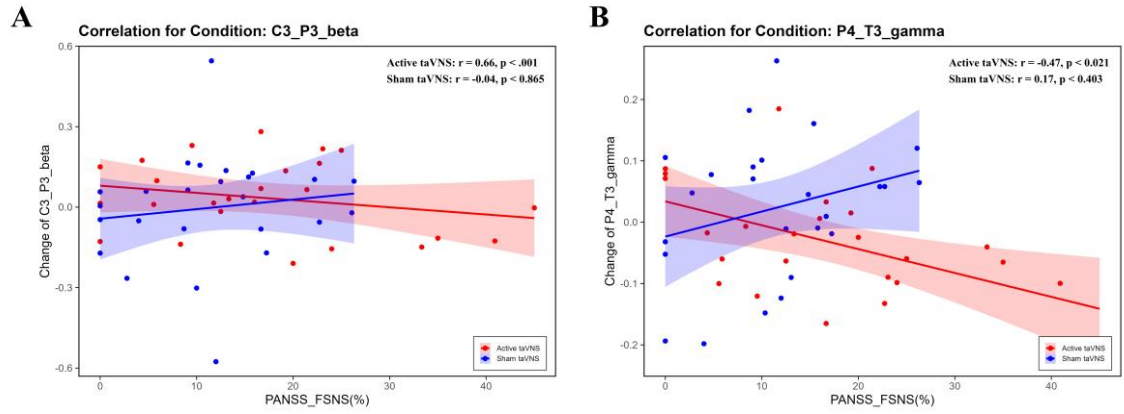

(A) Correlation between beta-band EEG coherence at F3-P3 and the improvement rate of PANSS-FSNS. (B) Correlation between gamma-band EEG coherence at P4-T3 and the improvement rate of PANSS-FSNS.

Table 1. Baseline demographic and clinical characteristics by treatment group.

| Characteristic | taVNS Group, Mean (SD) | | p-values |
|---|---|---|---|
| | Active (n = 25) | Sham (n = 25) | |
| Age, y | 50.84(8.46) | 49.60(8.80) | 0.618 |
| Male, No. (%) | 11(44) | 8(32) | 0.561 [b] |
| Years of study | 5.92(4.43) | 5.92(3.19) | >.999 |
| Duration of disease, y | 17.40(11.02) | 12.80(8.97) | 0.124 |
| Antipsychotic drug equivalent [a] | 14.45(6.98) | 13.88(6.25) | 0.748 |
| PANSS score | | | |
| Positive symptoms | 14.45(6.98) | 13.28(2.64) | 0.525 |
| Negative symptoms | 29.52(4.39) | 29.72(5.12) | 0.904 |
| General symptoms | 38.64(5.48) | 39.76(4.68) | 0.457 |
| Total symptoms | 82.04(9.91) | 82.76(7.32) | 0.784 |
| PANSS-FSNS score | 30.84(4.33) | 31.72(4.74) | 0.513 |
| PANSS-FSPS score | 9.52(2.58) | 8.96(2.35) | 0.448 |
| CDSS score | 2.20(2.80) | 2.52(3.27) | 0.783 |
| SANS score | 67.24(17.68) | 69.24(22.36) | 0.720 |
| CGI-S | 4.48(0.71) | 4.44(0.51) | 0.991 |

[a] Antipsychotic doses were converted to equivalents to olanzapine.
[b] Determined with Fisher exact test for count data.

Table 2. Averaged predictive performance of the five-fold nested cross-validation SVR model.

| SVR model | RMSE | MAE | $r$ [a] | $p$-values |
|---|---|---|---|---|
| Active taVNS: PANSS-FSNS outcome predictor | 0.06 | 0.05 | **0.87** | **<.001** |
| Sham taVNS: PANSS-FSNS outcome predictor | 0.11 | 0.09 | -0.36 | 0.079 |
| Active taVNS: PANSS-FSPS outcome predictor | 0.39 | 0.30 | 0.14 | 0.495 |

[a] Correlation coefficient between predicted and true observed values.

# Supplemental Online Content

**Methods**

**Table S1.** Comparison of the PANSS-FSNS outcome predictor's stratification effects on the PANSS-FSNS and other symptom trajectories.

**Fig. S1.** Workflow for the establishment of the predictive model.

**Fig. S2**. Results of the permutation test.

**Fig. S3**. Using EEG features as predictors, the features selected at least three times from the 5 CV iterations in prediction of symptom changes.

**References**

## Methods

**Study Subjects and Endpoints for Individualized Prediction**

We targeted TRS patients primarily presenting with negative symptoms. Patients with TRS was defined according to the 2016 TRRIP working group consensus as: insufficient symptom remission after treatment with at least two different antipsychotics at adequate doses (≥600 mg/day chlorpromazine equivalents), for an adequate duration (≥6 weeks per treatment), and with sufficient adherence (actual dose taken ≥80% of the prescribed dose), with insufficient symptom improvement (defined as <20% improvement as assessed by standardized rating scales)(Howes et al., 2017).

Exclusion criteria included: a PANSS Positive Symptom Factor Score (PANSS-FSPS) above 19; a score of ≥ 4 on two or more of the five PANSS-FSPS items (P1 delusions, P3 hallucinatory behavior, P5 grandiosity, P6 suspiciousness/persecution, G9 unusual thought content)(Leucht et al., 2019); a total score of >14 on the Calgary Depression Scale for Schizophrenia (CDSS)(Addington et al., 1993); insufficient education, inability to effectively communicate, or inability to complete study assessments; active suicidal or violent behavior as assessed by a professional psychiatrist; history or presence of epilepsy or other neurological disorders; severe hepatic, renal, cardiovascular, cerebrovascular, or hematological diseases; significant organic brain lesions, including traumatic brain injury or known structural brain abnormalities; diagnosis of comorbid major depressive disorder, anxiety disorder, obsessive-compulsive disorder, or other serious mental disorders; history of taVNS treatment within the last six months; auricular deformities or any other auricular disorders; pregnancy or breastfeeding, and those with fertility requirements or unable to use effective contraception within the past six months.

**EEG Data Acquisition and Processing**

**EEG preprocessing.** The preprocessing steps were as follows: (1) Applying a high-pass filter at 0.5 Hz and a low-pass filter at 40 Hz to process the data; (2) Using a notch filter (49-51 Hz) to eliminate 50 Hz power line noise; (3) Manually inspecting bad channels and performing bad channel interpolation using spherical splines, with no differences in the number of interpolated channels between groups (number of interpolations: baseline: active taVNS 0.68 (1.05), sham taVNS 0.80 (0.98); post-treatment: active taVNS 1.24 (1.21), sham taVNS 1.29 (1.31)); (4) Manually rejecting EEG segments contaminated by strong muscle artifacts or those with amplitudes exceeding ±100 μV at any electrode; (5) Removing artifacts, including ocular, muscle, and cardiac artifacts, using Independent Component Analysis(Jung et al., n.d.); (6) Re-referencing the EEG data to the average reference. The artifact-free preprocessed data were used for subsequent analysis, and there was no significant difference in the average duration of EEG data between the two groups (baseline: active taVNS 285.02s (87.89s), sham taVNS 265.05s (89.81s); post-treatment: active taVNS group 307.08s (84.43s), active taVNS 310.96s (80.06s)).

**EEG power.** The Welch method was used to perform fast Fourier transform on the preprocessed EEG data(Welch, 1967). Specifically, a moving window of 1000 points with a 50% overlap was used to obtain a power spectral density estimate in the range of 0.5-40 Hz (with a frequency resolution of 0.5 Hz). Mean power values were extracted from each electrode as well as from predefined brain regions, including the prefrontal (FP: Fp1, Fp2), frontal (F: F7, F3, F4, F8), central (C: C3, C4), parietal (P: P7, P3), and temporal (T: T3, T4, T5, T6) regions, across delta (0.5-4 Hz), theta (4-8 Hz), alpha (8-13 Hz), beta (13-30 Hz), gamma (30-40 Hz) and broadband (0.5-40 Hz) frequency bands.

**EEG coherence.** Coherence represents the normalized covariance of two time series in the frequency domain. Mathematically, the coherence function $Cxy(f)$ between signals x and y at frequency f is obtained by normalizing the cross-spectrum, as shown below:

$$Cxy(f) = \frac{|Sxy(f)|^2}{Sxx(f) \cdot Syy(f)}$$

Where Sxy(f) is the cross-spectrum, Sxx(f) and Syy(f) are the power spectra of each electrode, and Cxy(f) is the coherence between the two signals at frequency f. The coherence between electrode pairs in a specific frequency band is defined as the cross-spectral power between sites normalized by the square root of the product of the power at each site within the band(Grosse et al., 2002; R, 1987). EEG coherence was calculated using the mscohere function in MATLAB across all electrode pairs for the delta, theta, alpha, beta, gamma and broadband frequency bands.

**Dynamic functional connectivity.** A dynamic functional connectivity approach was used to characterize the spatiotemporal properties of brain states at the individual level. First, the EEG data were segmented using a sliding window method (window length: 4 s; step size: 1 s), and delta, theta, alpha, beta, and broadband components were extracted for each channel within each window. Phase information was then obtained via Hilbert transform, and weighted phase lag index (wPLI) matrices were computed across electrode pairs within each window. The wPLI matrices corresponding to each intervention group were aggregated and subjected to k-means clustering, yielding four prototypical brain states, each representing a stable group-level functional connectivity pattern(Chen et al., 2024). Using these templates, each wPLI matrix at the individual level was assigned to the most similar brain state, thereby generating a temporal sequence of brain states. Finally, dynamic features were extracted at the individual level, including the proportion, mean dwell time, and number of transitions of the four brain states, to quantify the temporal variability of brain dynamics.

**Establishment of a predictive model**

**Data Preprocessing**

Initially, missing values were addressed using mean imputation(Rubin, n.d.) to avoid data incompleteness. Subsequently, the interquartile range (IQR) method was applied to detect and correct outliers(Rubin, n.d.). For each continuous variable, the first quartile (Q1) and third quartile (Q3) were calculated, and the IQR was defined as:

$$IQR = Q3 - Q1$$

$$lower\ bound = Q1 - 1.5 \times IQR, \quad upper\ bound = Q3 + 1.5 \times IQR$$

Values outside these bounds were replaced with the mean of the corresponding variable. Finally, all features were standardized using Z-score normalization to eliminate scale differences across variables.

**Feature Selection**

To mitigate the impact of high-dimensional features on the predictive model, we employed an elastic net-based feature selection method(Guyon and Elisseeff, n.d.; Zou and Hastie, 2005). This method combines the sparsity of L1 regularization (Lasso) with the stability of L2 regularization (Ridge)(Hoerl and Kennard, n.d.; Tibshirani, 1996). effectively reducing multicollinearity-related interference on model performance while retaining key predictive features. We first performed five-fold cross-validation on the feature set and independently applied elastic net regression on the training set of each fold to select the most predictive features. In this process, to further assess the stability and robustness of feature selection, leave-one-out cross-validation (LOOCV) was applied within each training fold. During training, the indices of features selected by the elastic net model in each iteration were recorded. The objective function of the elastic net regression is as follows:

$$\min_{\beta} \left\{ \sum_{i=1}^{n} (y_i - x_i\beta)^2 + \alpha[(1-\lambda)\|\beta\|_2^2 + \lambda\|\beta\|_1] \right\}$$

In the objective function above, α is a hyperparameter that controls the overall strength of regularization,

while λ determines the relative contribution of L1 and L2 regularization. When λ is set to 1, the model simplifies to Lasso regression, applying only L1 regularization. Conversely, when λ is set to 0, the model reduces to Ridge regression, applying only L2 regularization. In the present study, α was set to 0.01 to ensure a moderate level of regularization, and λ was set to 0.8 to retain the primary characteristics of L1 regularization while simultaneously reducing instability caused by high-dimensional data.

During five-fold cross-validation, we recorded the selection frequency of each feature across folds. Features selected in at least three out of five folds were retained as the final subset. This strategy enhances the robustness of feature selection, reduces variability due to data partitioning, and ensures better model generalizability. The resulting feature subset was used for subsequent model training and optimization.

**Prediction model evaluation**

After determining the preliminary feature subset, we further optimized the feature combination using a forward stepwise regression strategy(Bendel and Afifi, 1977). Stepwise regression is a progressive feature selection method that begins with an empty model and iteratively introduces one feature at a time, evaluating its contribution to predictive performance. This approach reduces the risk of overfitting and ensures that the final model achieves optimal predictive accuracy with the minimal number of features. Specifically, we initiated the process with an empty model, subsequently adding one feature at each step, and trained the model using SVR with a linear kernel for evaluation.

To ensure the stability of selected features and the generalizability of the model, we employed a nested cross-validation scheme, where LOOCV was embedded within the five-fold cross-validation framework to objectively evaluate the predictive ability of different feature combinations(Varma and Simon, 2006). In each LOOCV iteration, the Pearson correlation coefficient ($r$) and corresponding $p$-value between the predicted and observed scores were calculated. The mean $r$-value across the five folds was used to quantify the overall predictive performance of the current feature set. This strategy enabled dynamic monitoring of model performance as features were incrementally added, thereby facilitating the identification of the optimal feature subset.

During stepwise regression, grid search was simultaneously applied to optimize the SVR hyperparameters $C$ ($10^{-2}$-$10^{3}$, logarithmic scale) and $\varepsilon$ (0.01-1.0, linear scale)(Bergstra and Bengio, n.d.). Hyperparameter tuning was conducted within the inner five-fold cross-validation to avoid data leakage, and the configuration yielding the highest mean r-value was retained. The feature subset achieving the highest mean r-value across folds was selected as the final set, which was further evaluated in the outer test folds to validate its generalizability.

# Table and Figure

Table S1. Comparison of the PANSS-FSNS outcome predictor's stratification effects on the PANSS-FSNS and other symptom trajectories.

| Outcome | Active Group, Mean (SD) | | Mixed effects model analysis | |
|---|---|---|---|---|
| | Predicted PANSS-FSNS Responders (n = 9) | Predicted PANSS-FSNS Non-Responders (n = 16) | F | P Value |
| Change in PANSS-FSNS | | | | |
| Week 2 (treatment endpoint) | -6.22(2.05) | -2.63(1.86) | 7.31 | **<.001** |
| Week 3 (follow-up period) | -7.44(1.88) | -4.31(2.18) | | |
| Week 4 (follow-up endpoint) | -8.89 (1.62) | -5.69 (3.14) | | |
| Change in PANSS-FSPS | | | | |
| Week 2 (treatment endpoint) | -1.00(1.12) | -2.13(2.06) | 1.21 | 0.314 |
| Week 3 (follow-up period) | -1.67(1.66) | -2.69 (2.15) | | |
| Week 4 (follow-up endpoint) | -2.11(2.32) | -3.19(2.32) | | |
| Change in PANSS negative symptoms subscale | | | | |
| Week 2 (treatment endpoint) | -5.22(1.39) | -2.63(2.45) | 3.99 | **0.011** |
| Week 3 (follow-up period) | -6.33(1.41) | -4.13 (2.87) | | |
| Week 4 (follow-up endpoint) | -7.22(1.09) | -5.94 (2.98) | | |
| Change in PANSS positive symptoms subscale | | | | |
| Week 2 (treatment endpoint) | -2.89(2.08) | -2.38 (2.53) | 0.66 | 0.582 |
| Week 3 (follow-up period) | -3.44 (2.83) | -3.50 (2.80) | | |
| Week 4 (follow-up endpoint) | -4.00 (3.39) | -4.81 (3.08) | | |

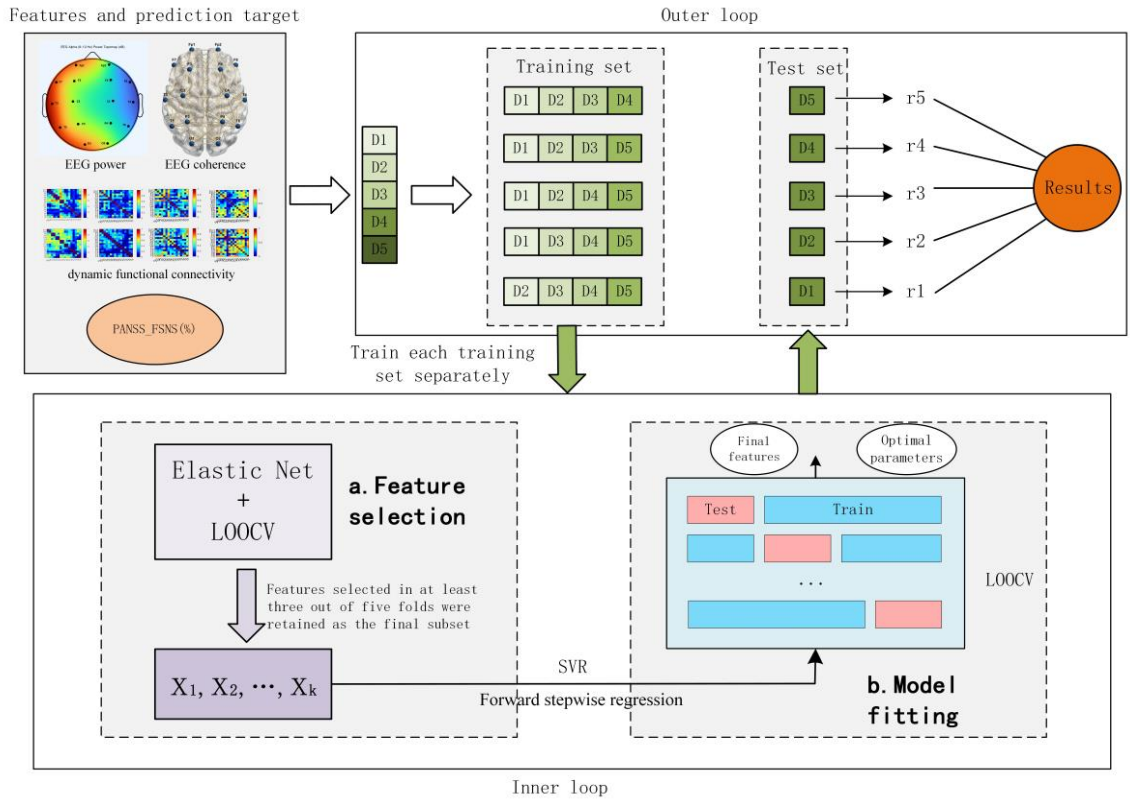

**Fig. S1.** Workflow for the establishment of the predictive model.

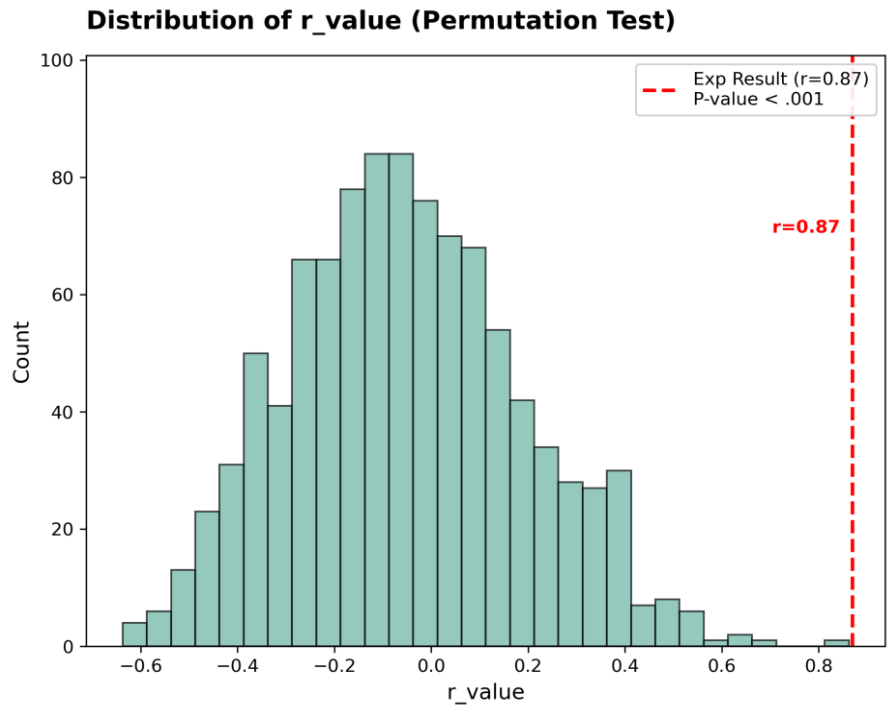

**Fig. S2.** Results of the permutation test.

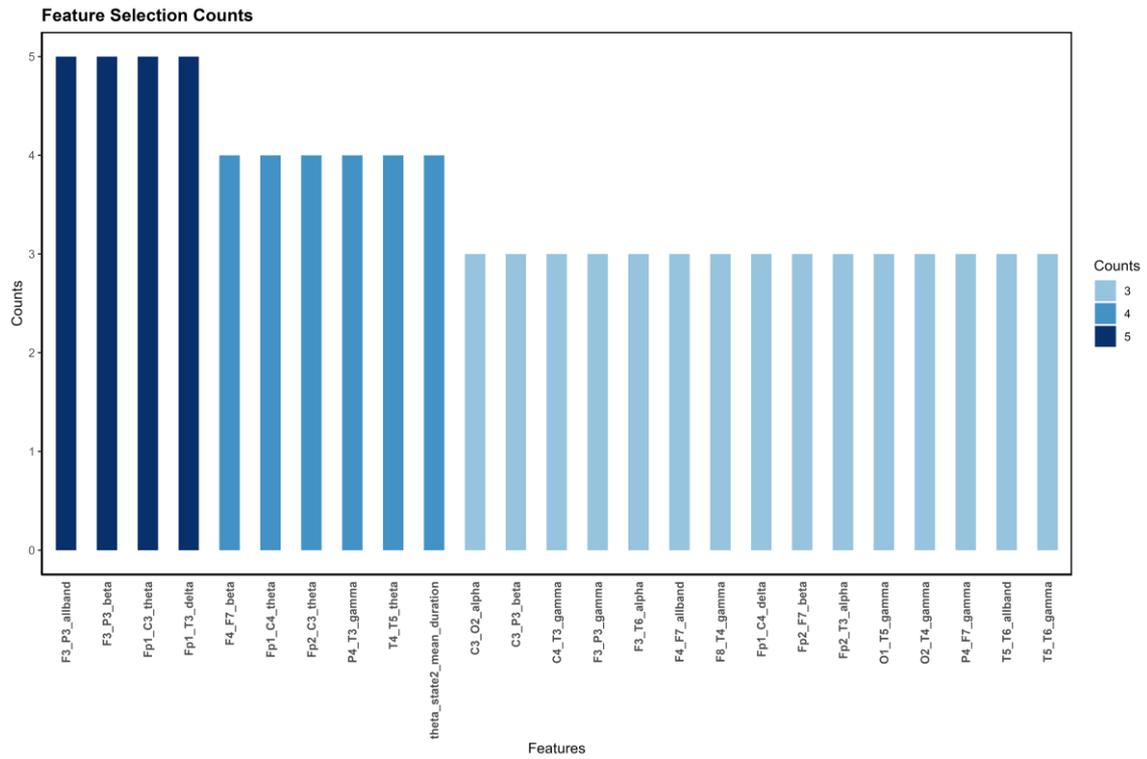

**Fig. S3.** Using EEG features as predictors, the features selected at least three times from the 5 CV iterations in prediction of symptom changes.